\documentclass[12pt]{article}
\usepackage{lscape}
\usepackage[utf8]{inputenc}
\usepackage[T1]{fontenc}
\usepackage{lmodern}
\usepackage{subcaption}
\usepackage[english]{babel} 
\usepackage[babel]{csquotes}
\usepackage{booktabs,tabularx}
  \newcolumntype{Y}{>{\centering\arraybackslash}X}
\usepackage{array,longtable,multirow}
\usepackage{graphicx}
\usepackage[font=small,labelfont=bf,singlelinecheck=false]{caption} 
\usepackage{amsmath}
\usepackage{amssymb}
\usepackage{amsmath}
\usepackage{float}
\usepackage{xcolor}
\usepackage{algorithm}
\usepackage{algpseudocode}
\usepackage[doublespacing]{setspace} 
\usepackage[round]{natbib}
\usepackage[breaklinks, linktocpage, bookmarksnumbered=true, colorlinks=true, linkcolor=cyan, citecolor=cyan, urlcolor=cyan, pdftitle={tyranny},  pdfauthor={Matteo Borrotti}]{hyperref} 
\usepackage{geometry}
\usepackage{rotating}
\usepackage{subfiles}
\usepackage{tikz}
\usepackage{lscape}
\usepackage{pdflscape} 
\usepackage{afterpage}
\usepackage{multicol}		
\usepackage{authblk}

\usepackage{fancyhdr}
\providecommand{\keywords}[1]
{
  \small	
  \textbf{\textit{Keywords---}} #1
}
	
\definecolor{chameleongreen1}{RGB}{98,189,25}
\renewcommand*{\thefootnote}{\fnsymbol{footnote}}

\fancypagestyle{lscapedplain}{%
  \fancyhf{}	
  
	\fancyfoot{%
    \tikz[overlay]
      \node[outer sep=2cm,above,rotate=90] at (current page.east) {\thepage};}
}

\usepackage{titlesec}

\titleformat*{\section}{\large\bfseries}
\titleformat*{\subsection}{\normalsize\bfseries}
\titleformat*{\subsubsection}{\normalsize\bfseries}
\titleformat*{\paragraph}{\normalsize\bfseries}
\titleformat*{\subparagraph}{\normalsize\bfseries}

\DeclareMathOperator*{\argmax}{argmax}
\large

\makeatletter
\def\@footnotecolor{red}
\define@key{Hyp}{footnotecolor}{%
 \HyColor@HyperrefColor{#1}\@footnotecolor%
}
\def\@footnotemark{%
    \leavevmode
    \ifhmode\edef\@x@sf{\the\spacefactor}\nobreak\fi
    \stepcounter{Hfootnote}%
    \global\let\Hy@saved@currentHref\@currentHref
    \hyper@makecurrent{Hfootnote}%
    \global\let\Hy@footnote@currentHref\@currentHref
    \global\let\@currentHref\Hy@saved@currentHref
    \hyper@linkstart{footnote}{\Hy@footnote@currentHref}%
    \@makefnmark
    \hyper@linkend
    \ifhmode\spacefactor\@x@sf\fi
    \relax
  }%
\makeatother
\hypersetup{footnotecolor=black}

\title{Dealing with uncertainty: balancing exploration and exploitation in deep recurrent reinforcement learning}

\author[1]{Valentina Zangirolami}
\author[1]{Matteo Borrotti}

\affil[1]{University of Milano-Bicocca, Milan, Italy}

\date{}

\begin{document}

\maketitle 

\begin{abstract}
\noindent Incomplete knowledge of the environment leads an agent to make decisions under uncertainty. One of the major dilemmas in Reinforcement Learning (RL) where an autonomous agent has to balance two contrasting needs in making its decisions is: exploiting the current knowledge of the environment to maximize the cumulative reward as well as exploring actions that allow improving the knowledge of the environment, hopefully leading to higher reward values (exploration-exploitation trade-off). Concurrently, another relevant issue regards the full observability of the states, which may not be assumed in all applications. For instance, when 2D images are considered as input in an RL approach used for finding the best actions within a 3D simulation environment. In this work, we address these issues by deploying and testing several techniques to balance exploration and exploitation trade-off on partially observable systems for predicting steering wheels in autonomous driving scenarios. More precisely, the final aim is to investigate the effects of using both adaptive and deterministic exploration strategies coupled with a Deep Recurrent Q-Network. Additionally, we adapted and evaluated the impact of a modified quadratic loss function to improve the learning phase of the underlying Convolutional Recurrent Neural Network. We show that adaptive methods better approximate the trade-off between exploration and exploitation and, in general, Softmax and Max-Boltzmann strategies outperform $\epsilon$-greedy techniques.
\end{abstract}


\setcounter{footnote}{0}
\renewcommand*{\thefootnote}{\arabic{footnote}}

\section{Introduction}\label{sec:intro}

Reinforcement learning (RL) is a core topic in machine learning and is concerned with sequential decision-making in an uncertain environment. Two key concepts in RL are exploration, which consists of learning via interactions with an unknown environment, and exploitation, which consists of optimizing the objective function given accumulated information. In such a scenario, an RL algorithm repeatedly makes decisions to maximize its rewards, the so-called exploitation; the RL algorithm, however, has only limited knowledge about the process of generating the rewards. Thus, occasionally, the algorithm might decide to perform exploration which improves the knowledge about the reward generating process, but which is not necessarily maximizing the current reward \cite{auer2002}. Exploitation can be studied using the (stochastic) control theory, while exploration relies on the theory of statistical learning. These two concepts are complementary but opposite: exploration leads to the maximization of the gain in the long run at the risk of losing short-term reward, while exploitation maximizes the short-term gain at the price of losing the gain over the long run. A careful trade-off between these two objectives is important to the success of any learner.

The main motivation of this work is to address the issues of partial observability of states together with the exploration-exploitation trade-off through deterministic and stochastic strategies. The final aim is to provide a comprehensive analysis of reinforcement learning, with a focus on deep recurrent reinforcement learning from the point of view of the previously mentioned trade-off. In summary, the main contributions of this work can be summarized as follows:
\begin{itemize}
    \item[(1)] Refine the quadratic loss function proposed by \citet{DRQN_2} to align with the Bootstrapped Random Update sampling technique. While \citet{DRQN_2} originally developed this strategy for Bootstrapped Sequential Update sampling technique. The introduction of this strategy to deep recurrent reinforcement learning leads to a speed-up of the algorithm convergence.
    \item[(2)] Conduct a comprehensive deep analysis and comparison of several exploration strategies on partially observable systems. Specifically, we evaluated several $\epsilon$-greedy and Softmax approaches, incorporating Bayesian perspectives and assessing the agent's uncertainty to adjust the $\epsilon$ probability dynamically.  Originally, these methods were proposed by \citet{VDBE}, \citet{VDBE_softmax}, and \citet{BMC} for fully observable system states, we successfully adapted all strategies to the recurrent framework.
    \item[(3)] Modify the VDBE method to suit its adaptation and integration into the Deep Recurrent Q-learning and, more broadly, Deep Q-Learning scenarios. Indeed, we consider the difference among estimated Q-values with different estimated weights of neural networks (related to previous and current updates of the agent).
    \item[(4)] Implement a simulation study to predict the steering wheel angle in autonomous driving systems. Within this application, we examined the impact of the exploration strategies in partially observable systems for balancing exploration. To address the challenges posed by partially observable states, we proposed Deep Double Dueling Recurrent Q-Learning with a modified loss to enhance the performance of collision avoidance while keeping the speed constant, evaluating solely the steering wheel of the vehicle.
\end{itemize}

This paper is organized as follows. Section \ref{sec:soa} describes related work based on exploration strategies, recurrent models of DRL and autonomous driving. Section \ref{sec: background} provides a comprehensive overview of reinforcement learning, highlighting key issues and introducing deep recurrent reinforcement learning. Section \ref{sec: drrl} delves into the details of Deep Recurrent Q-learning model tailored for self-driving cars specifying our loss proposal for Boostrapped Random updates. Section \ref{sec:expl_str} elucidates the exploration strategies compared in the experiments emphasizing their theoretical aspects within a partial observable framework. Section \ref{sec:exp} outlines the experimental settings, encompassing the use of AirSim simulator, the parameters of exploration strategies, and DRL model. Additionally, it shows the results obtained from the training and test set, along with a deep analysis of the impacts of exploration strategies. Section \ref{sec:conclusion} concludes this paper with final considerations and outlines for future works.

\section{Related literature}\label{sec:soa}

This work is related to three major research fields, namely exploration strategies, function gradient approximation for RL and autonomous driving.

\subsection{Exploration strategies}
In the literature, many different exploration methods for RL processes are based on a discrete action space, in which Softmax and $\epsilon$-greedy are the most popular. \citet{Ortiz_epsgreedy} considered a deterministic $\epsilon$-greedy strategy with linear decreasing using four different models, including Q-learning. This kind of $\epsilon$-greedy method is significantly simplified, where $\epsilon$ is the inverse of the step number without defining the lower bound to be reached in the final step. \citet{BMC} and \citet{VDBE} proposed two novel adaptive methods for $\epsilon$-greedy on full-observable domain and used training information to update $\epsilon$ probability. \citet{VDBE} compared constant $\epsilon$-greedy and softmax with an adaptive $\epsilon$-greedy strategy based on Q-value differences as a measure of uncertainty. \citet{BMC} used Bayesian Inference to estimate the $\epsilon$ value with theoretical convergence guarantee and compare it with deterministic and adaptive $\epsilon$-greedy approaches. However, both papers conducted a limited comparison of strategies. \citet{BMC} compare only $\epsilon$-greedy strategies while \citet{VDBE} only includes constant $\epsilon$-greedy while Softmax. \citet{VDBE_softmax} proposed an extension of Max-Boltzmann Exploration in a full-observable states domain which combines Softmax and value-based adaptive $\epsilon$-greedy. \citet{VDBE_softmax} compared this novel strategy with constant $\epsilon$-greedy, VDBE, and Softmax regardless of decreasing $\epsilon$-greedy and other adaptive strategies. \citet{CruzeWupppen} studied the effect of many exploration strategies. Among others, the baseline form of $\epsilon$-greedy and Softmax with the extensions proposed by \citet{VDBE_softmax} and \citet{VDBE}. Our work extends the previous analysis by investigating the impact of other techniques such as Max-Boltzmann Exploration and Decreasing $\epsilon$-greedy strategy, where the decreasing linear equations assume changes in slope depending on the progress in the training learning. Furthermore, we analysed all these methods in order to evaluate the balance of exploration-exploitation trade-off on a partially observable system.

\subsection{Function gradient-based RL}

The success of RL in decision-making has recently enticed researchers to apply Deep Q-Learning (DQL) methods on video games and autonomous driving tasks, which all require Convolutional Neural networks to approximate value functions with image-based states. In literature, there are several models used to estimate Q-values in discrete control systems which could be based on Full and Partial observability of the states. \citet{KIM2022116742} suggested the extension of Deep Q-Network (DQN) approach, like Double Deep Q-Network (DDQN) and Double Dueling Deep Q-Network (D3QN), to find the optimal action avoiding obstacles on autonomous drone. \citet{KIM2022116742} showed that DDQN and D3QN overcome the performance of the baseline DQN structure, in which D3QN learns better policies than other methods. Several works \cite{DQLeDQP, computers11030041, Zhang_dqn, Min_dqn, HAN2019113708} evaluated Deep Q-Learning in several RL applications. However, DQN and its extensions assume the full observability of the states that fall in the real world and, in general, 3D environment. The real-world environment is characterised by uncertainty whereas a system based on full observability fails to capture the true dynamics as there might be noisy sensors, missing information about the state, or outside interferences. \citet{DRQN} proposed two main sampling methods of Deep Recurrent Q-Learning (DRQL) and they used Bootstrapped Random Updates from memory replay for Neural Network weight optimization. Moreover, \citet{DRQN} compared DQN and Deep Recurrent Q-Network (DRQN) methods on two game environments, where DRQN performed both well and poorly. In fact, \citet{DRQN} showed how DRQN updates might trigger some problems with the learning of functions which could be reflected in the final performance. Other papers \cite{DRQNvsDQN, Zeng_drqn, JiajunOu, Xu_drqn} compared Deep Q-Network and Deep Recurrent Q-Network in different fields to evaluate the performance on a partially observable domain. However, most of these studies did not achieve good results when systems were based on hidden states. \citet{DRQN_2} proposed a novel method for Deep Recurrent Q-Network so as to overcome challenges during the agent's update. Consistently, \citet{DRQN_2} proposed a technique based on an error mask in the optimization phase in order to update neural network weights with enough history of observations. \citet{DRQN_2} tested their techniques coupled with Bootstrapped Sequential Update as a sampling strategy. In this work, we extend the contribution of \citet{DRQN_2} to Bootstrapped Random Update sampling strategy.

\subsection{Autonomous driving}

The use of Convolutional Neural Network (CNN) in Autonomous Driving has been a topic of great interest in recent years. Some works have applied Deep Reinforcement Learning in order to automate vehicle and process images with more scalability. Several researchers studied autonomous driving tasks by using simulators that generate iterative images and perform actions in real time. \citet{Wu_ad} and \citet{Santara_2021} considered TORCS simulator platform to build DRL frameworks. Generally, TORCS is used to simulate a car racing environment, which is useful for managing and training the agent in situations where other cars are present. However, a car racing environment does not include several real-world peculiarities like parked cars, static obstacles, animals, and vegetations. 
\citet{Xiao_ad} and \citet{Michelmore_ad} show the results obtained by two different DRL frameworks using CARLA simulator. In comparison to TORCS, CARLA provides environments that are visually closer to reality. AirSim and CARLA are very similar, which is the reason why \citet{Pilz_ad} assigned positive ratings to these two simulators in terms of interface compatibility, access to ego vehicle data, access to nonego vehicle data, access to pedestrian data, detail and variety of sensors, detail of the rendered graphics, detail of the physics engine and cost efficiency. In the literature, there are some works on DRL for autonomous driving. \citet{DQN_conference} compared DDQN and D3QN models with transfer learning techniques via decreasing $\epsilon$-greedy in order to evaluate collision avoidance performance assuming a discrete action space of the steering angle. \citet{drqn_ad_deteps} proposed a DRQN model with descending $\epsilon$-greedy to control car's steering angle and speed, as \citet{Liao_ad} compared DQN and DDQN models for the same objective. Our study aims to overcome these papers by considering several exploration strategies on a partially observable system, which proves a central issue in dealing with 3D environments. Moreover, previous works only considered deterministic strategies for balancing exploration, which may not have been efficient in approximating agent uncertainty because of a $\epsilon$ exogenous.

\section{Background and preliminaries} \label{sec: background}

In this section we delve into the concept of \emph{exploration} within reinforcement learning, presenting prior frameworks addressing similar challenges. We subsequently introduce fundamental notions and theoretical background of reinforcement learning, emphasizing its application in partially observable environments and exploring the realm of deep recurrent reinforcement learning.

\subsection{Multi-Armed Bandits and Contextual Bandits}

The multi-armed bandit problem (\citet{mab_1}, \citet{mab_gitting_index}, \citet{bubeck2012regret}) is the simplest framework that encompasses the key issue of \emph{balancing exploration and exploitation}. In the aim of maximizing the cumulative discounted future rewards, the current decision about the action to be taken revolves around a dilemma: yielding the highest reward (\emph{exploitation}) or exploring new actions (\emph{exploration}) thereby gaining additional knowledge for improving future decisions.

The stochastic multi-armed bandit problem could be represented by a set of actions $\mathcal{A} = \{1, \ldots, K\}$, an agent which should select one action (or arm) at every time step $t$ until a finite horizon $t=1, \ldots, T$ and a sequence $r_{k, 1}, \ldots, r_{k, T}$ of unknown rewards associated with each arm $k$. Hence, at each time step $t$ an agent, after choosing an action $a_t$, receives a reward $r_{a_t, t}$ which is drawn from an unknown probability distribution $R_{a_t}$ (independently from the past). The effectiveness of the agent's strategy is evaluated by comparing it with the optimal action in expectation, with the aim of minimizing the pseudo-regret. The stochastic contextual bandit enriches the multi-armed bandit framework by incorporating context sets $\mathcal{S}$. The new objective of an agent is to identify the optimal policy within a class of policies that maps the context to the action.

\subsection{Reinforcement Learning}

Multi-armed bandit and contextual bandit frameworks can be seen as particular cases of a full reinforcement learning problem. RL (\citet{rl_kaelbling}, \citet{rl_sutton}) is formulated as (completely observed) Markov Decision Process (MDP), where current actions can affect the next states. A MDP describe a decision-making process and it can be defined as a tuple $(\mathcal{S}, \mathcal{A}, \mathcal{T}, \mathcal{R}, \gamma)$, where $\mathcal{S}$ is a set of states which contain information about the environment for each time step, $\mathcal{A}$ is a set of potential actions to be taken based on the states received by an agent and $\mathcal{R}$ is a set of rewards. $T$ is a set of the transition probabilities $T_{sas'} = \mathbb{P}(S_t= s'| S_{t-1}=s, A_{t-1}=a)$ describing the dynamics of the environment, and $\gamma \in (0,1)$ is a discount factor which is used to compute discounted return. 

The most popular RL methods, such as Deep Q-learning, involve action-value functions $Q_\pi(s,a) = \mathbb{E}_\pi[\sum_{k=0}^{\infty} \gamma^k R_{t+k+1} | S_t =s, A_t =a ]$ to encode the policy $\pi$, which maps the states to the probabilities of selecting each action $a \in \mathcal{A}$. To solve the RL problem, the aim is to find a class of optimal policies that corresponds to finding the optimal action-value function $Q^*(s,a) = \max_{\pi} Q_\pi(s,a)$. Hence, an optimal policy can be reached by $\pi^*(a|s) \in \argmax_a Q^*(s,a)$.

\subsection{Exploration strategies}

Exploration strategies are employed to balance the exploration-exploitation trade-off in the above settings. The most popular methods are, among others, $\epsilon$-greedy, softmax, and Max-Boltzmann methods. In the case of Deep Q-learning (assuming full observable states), exploration strategies lead to exploring new actions rather than following a greedy policy, concerning the optimal action-value function, throughout the process. For instance, $\epsilon$-greedy techniques balance the exploration-exploitation trade-off through a probability $\epsilon \in (0,1)$ such that
\begin{gather}
    \pi_{\epsilon} (a|s) = \begin{cases}
        1- \epsilon_t + \frac{\epsilon_t}{|\mathcal{A}|}, & \mbox{if } a^*=\argmax_{a \in \mathcal{A}} Q(s,a)\\
        \frac{\epsilon_t}{|\mathcal{A}|}, & \mbox{otherwise}
    \end{cases} \mbox{ ,}
\end{gather}
where $t=1, \ldots, T$ refers to time steps. Softmax exploration encompasses Boltzmann distribution function such that
\begin{gather}
    \pi_B(a|s) = \frac{e^{\frac{Q(s,a)}{\tau}}}{ \sum_{l \in \mathcal{A}} e^{\frac{Q(s,l)}{\tau}}}\mbox{ ,}
\end{gather}
where $\tau$ regulates the trade-off between exploration and exploitation.

\subsection{Recurrent Reinforcement Learning}

Recurrent Reinforcement Learning is formulated as Partially Observable Markov Decision Process (POMDP), which deals with partially observable states instead of fully observable states as in MDP. POMDP (\citet{Hauskrecht_2000}) can be seen as a generalization of MDP, allowing to model and reason about the uncertainty on the current state of the system in sequential decision problems (\citet{JMLR:v12:ross11a}). It is defined as a tuple ($\mathcal{S}, \mathcal{Z}, \mathcal{O}, \mathcal{A}, \mathcal{T}, \mathcal{R}, \gamma$), where $\mathcal{Z}$ is a set of observations and $\mathcal{O}$ is the set of observation probabilities $O_{saz}=\mathbb{P}(Z_t=z|S_t=s, A_t=a)$. As in the previous case, Deep Q-learning encodes the policy by using Q-values which are estimated through a Deep Neural Network. Hence, the optimal policy can be reached by finding the optimal action-value function $Q^*(h, a)= \max_{\pi} Q_{\pi}(h,a)$, where $H_{t-1}=h$ is the previous information state.

\section{Deep Recurrent Q-Learning for autonomous driving} \label{sec: drrl}

\begin{figure}[ht]
	\begin{center}
	    \includegraphics[scale=1]{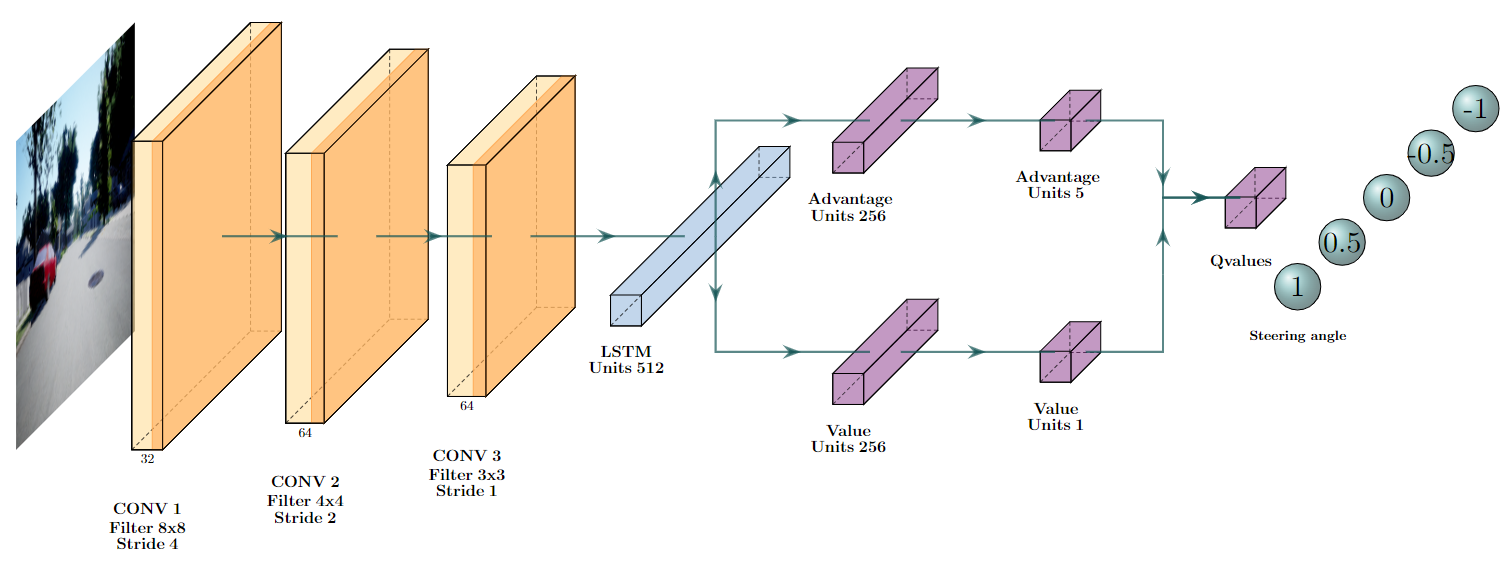}
	    \caption[DRQN]{\textbf{Convolutional Neural Network with LSTM layer.} Each pre-processed input image was processed by three convolutional layers and the resulting activations were performed through LSTM layer. Q-values were estimated by dividing the latter output into advantage and value fully-connected layers respectively and combining them together.}
	    \label{fig:ddrqn}
	\end{center}
\end{figure}

In this section, we outline the proposed approach, based on a deep recurrent reinforcement learning framework, as a solution to autonomously control the steering angle of cars. Concurrently, we compare several exploration strategies to achieve an improved balance of exploration-exploitation trade-off, intending to enhance obstacle avoidance performance (Section \ref{sec:expl_str}). We also enrich our proposed method by incorporating task-specific rewards and implementing pre-processing techniques on the states.
Figure \ref{fig:ddrqn} shows the agent's network, the state representation, and the action set which are involved in the reinforcement learning framework.
At each step, $t$, the state $s_t$ is processed by the agent's network (in Figure \ref{fig:ddrqn}) which is used to estimate Q-values for each action. The agent chooses the action $a_t$ which corresponds to the maximum of Q-values. Then, the reward value $r_t$ is computed, and performing the action $a_t$, the environment provides the next state $s_{t+1}$. At a specific moment of the process, the agent's network is updated with traces of experience samples in a Bootstrapped Random Update fashion (Section \ref{subsec:agentsarchitecture}).


\subsection{Observation space}

\begin{figure}[ht]
	\begin{center}
	    \includegraphics[scale=1.2]{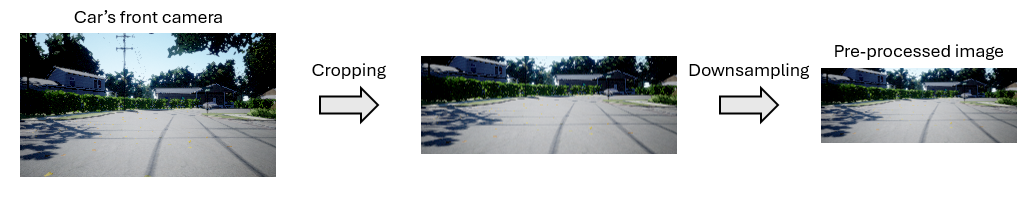}
	    \caption[Image]{\textbf{Pre-processing of car's front camera images.} The two-step pre-processing procedure is illustrated. Initially, each input image undergoes cropping at the top and bottom to concentrate the view of the road. Subsequently, downsampling is employed on the images before feeding them into the neural network.}
	    \label{fig:preprocdata}
	\end{center}
\end{figure}

In our framework, the observations are represented by images captured by the car's front camera, extracting information from the surrounding environment. At each time step, an image is collected, providing a snapshot of the current road conditions ahead of the car. To highlight the road and the related obstacles, each image has been pre-processed by using cropping techniques. Additionally, to streamline image processing and reduce dimensionality, we apply downsampling on the cropped images, effectively eliminating non-essential information (see Figure \ref{fig:preprocdata}). The resulting size corresponds to $200 \times 66$ pixels.

\subsection{Action space and Reward}
To avoid obstacles, we consider a set of discrete actions, each representing distinct values for the car's steering angle. The action set $\mathcal{A}= \{-1, -0.5, 0, 0.5, 1\}$ encompasses two levels of left steering, straight-ahead steering, and two levels of right steering. When an action is chosen, the car performs the related steering while keeping a constant speed.

We define a task-specific \emph{Reward} by incorporating a measure of the distance between the position of the car ($x_e$) and the centre of the roads ($x_r$), such that
\begin{gather}
    r(s, s', a)  = e^{-\beta \cdot ||x_e -x_r||}\mbox{ , } \label{eq:reward}
\end{gather}
where $\beta$ is a positive constant and $r \in [0,1]$ is the reward function. This formulation, proposed by \citet{DQN_conference}, and devised by \citet{Reward_definition}, is used to measure the goodness of steering angle actions, where the best action is achieved when the car is placed in the center of the road.

\subsection{Agent's architecture} \label{subsec:agentsarchitecture}

We consider the Double Dueling Deep Recurrent Q-Network architecture as the agent's network, where the weights are updated by using Bootstrapped Random Update. We propose to combine a modified loss function which leads to update weights only for the last observations of the episode trace (\citet{DRQN_2}) and the Boostrapped Random Updates. 

The agent's network combines convolutional and LSTM layers to estimate advantage $A(h_t, a;\theta, \alpha)$ and value function $V(h_t; \theta, \beta)$ for obtaining Q-value estimates (see Figure \ref{fig:ddrqn}) such that 
\begin{gather}
    Q(h_t,a; \theta)= V(h_t; \theta, \beta) + \bigl( A(h_t, a;\theta, \alpha) - \frac{1}{|A|}\cdot \sum_{a'} A(h_t, a';\theta, \alpha) \bigl) \mbox{ , }\label{eq:q_values}
\end{gather}
where $\theta$ are the weights of the main network. The weights of the main network are updated by using a modified quadratic loss function, which is defined as
\begin{gather}
    L(\theta_i) =\begin{cases} 0, &\mbox{if } i\le n_{err} \\
    E_{(o_t, a_t, r_t, o_{t+1})}\bigl[(y_i - Q(h_t, a_t; \theta_i)^2\bigl], &\mbox{otherwise} \end{cases}\mbox{ , } \label{eq:loss_function}
\end{gather}
where $i=1, \dots, T$ are the time steps of the LSTM layer. $\theta, \theta'$ are the weights of the main network and the target network, respectively. $n_{err}$ is the number of errors masked. The target $y_i$ are estimated using a double estimator, as $y_i = r_t + \gamma Q(h_{t+1}, \argmax_a Q(h_{t+1}, a; \theta_i); \theta_i')$. During the weights updating process, Bootstrapped Random Updates are used as a sampling method. The choice of this method requires the LSTM’s hidden state to be zeroed at the beginning of each update, and this peculiarity can make the learning of recurrent neural networks harder \cite{DRQN}. Consequently, to mitigate this challenge, only accurate gradients are propagated at each agent update through the network masking the first losses, to guarantee the update of the states with sufficient history \cite{DRQN_2}. The sample sequences are built by randomly drawing out episodes from the experience replay, followed by the definition of an experience trace. The length of this experience trace is regulated by a specific parameter, and the initial state is randomly selected from the sampled sequences (see Algorithm \ref{alg:bru}). Finally, the target network weights are updated by using the following rule
\begin{gather}
    \theta'= \theta \cdot \eta + \theta' \cdot (1- \eta)\mbox{ ,}
\end{gather}
where $\eta$ is a parameter that regulates the proportion of main network weights involved in the updated target network weights. 

\begin{algorithm}
\caption{Bootstrapped Random Update with masking of errors}\label{alg:bru}
\begin{algorithmic}
\State Buffer Size \emph{n}, Trace Length \emph{t}, Batch Size \emph{b}, Loss Function \emph{L}
\State Neural Network (Main) \textbf{MN}, Neural Network (Target) \textbf{TN}
\State Replay Memory $\mathcal{M}= \{ \mathcal{S}_{t}, \mathcal{A}, \mathcal{S}_{t+1}, \mathcal{R}, \mathcal{T}\}$
\If{$length(\mathcal{M}) \ge \frac{n}{2}$} \Comment{Update starts}
    \State sample\_episode = random\_sample($\mathcal{S}_t$, n\_sample = $b$) \Comment{Indicator of episodes}
\EndIf
\For{i in sample\_episode} \Comment{Sampling}
\State sample\_step = random\_sample($\mathcal{S}_t$[i,], n\_sample = 1)
\State trace($\mathcal{S}_{t}$)= $\mathcal{S}_t$[i, sample\_step : sample\_step+t]  \Comment{For each element of $\mathcal{M}$}
\EndFor
\State $main\_Q$ = \textbf{MN}(trace($\mathcal{S}_{t+1}$))  \Comment{According to \eqref{eq:q_values}}
\State $target\_Q$ = \textbf{TN}(trace($\mathcal{S}_{t+1}$)) \Comment{According to \eqref{eq:q_values}}
\State $double\_Q = target\_Q[, \argmax_a main\_Q]$ \Comment{$a \in \mathcal{A}$}
\State $Y= trace(\mathcal{R}) + \gamma \cdot double\_Q \cdot (1 - trace(\mathcal{T}))$
\State $Q= \max_a \textbf{MN}(trace(\mathcal{S}_{t}))$
\State $L_j = \frac{\sum_{i=1}^t w_i \cdot (Y_{j,i} - Q_{j,i})^2}{t}$ \Comment{$w_i \in \{ 0,1\}$ with \eqref{eq:loss_function} condition; $\forall j \in \{1, \ldots, b\}$}
\end{algorithmic}
\end{algorithm}

\section{Exploration strategies for recurrent learning} \label{sec:expl_str}
The behavioral policies of our agents are contingent upon the selected exploration strategy. Together with the proposed method for autonomous driving (see Section \ref{subsec:agentsarchitecture}), we conduct a comparative analysis of various exploration strategies, encompassing both deterministic and adaptive approaches. Therefore, we present a theoretical overview of these strategies within deep recurrent reinforcement learning. We consider many $\epsilon$-greedy and softmax methods, as well as their combinations. Within a partial observable process, the $\epsilon$-greedy policy can be defined as
\begin{gather}
    a_t=\begin{cases}\argmax_a Q_t(h_t,a), & \mbox{with probability } 1-\epsilon \\ \mbox{any action(a)}, & \mbox{with probability } \epsilon \end{cases} \mbox{ .}\label{eq:eps_greedy}
\end{gather}
The $\epsilon$-greedy strategy balances exploration and exploitation by employing $\epsilon$, which represents the exploration probability at each step. During the exploration phase, actions are selected randomly. Hence, how to adjust the $\epsilon$ probability poses one of the greatest challenges of RL, since it measures the degree to which the RL process should explore. The $\epsilon$-greedy is widely used and can be adapted in various forms depending on how $\epsilon$ is determined, thus considering deterministic or stochastic approaches. However, a notable drawback of this strategy concerns the random policy during the exploration phase. Although exploration is intended to facilitate the discovery of new actions for optimal decision-making in the future, the use of a random policy introduces the risk of obtaining significantly worse actions. To address this issue, we compare $\epsilon$-greedy strategies with Softmax and Max-Boltzmann Exploration methods.

Softmax is based on a different policy in which actions are drawn out from Boltzmann distribution. In the case of Deep Recurrent Q-Learning, we can define the Softmax policy as
\begin{gather}
    \pi_{B}(a|h_t)=\frac{e^{\frac{Q(h_t,a)}{\kappa}}}{\sum_b e^{\frac{Q(h_t,b)}{\kappa}}} \mbox{ .}\label{eq:softmax}
\end{gather}
Despite this, the Softmax method entails continuous exploration throughout the process, wherein the degree of separability from a completely random policy is regulated by $\kappa$, known as the temperature, not leaving the possibility of integrating the greedy actions.

\subsection{Deterministic \texorpdfstring{$\epsilon$}{ε}-greedy}
Using a deterministic strategy for $\epsilon$-greedy, the evolution of the $\epsilon$ probability is fixed and exogenously defined independently of the ongoing process. We explore two deterministic approaches, including the simplest form known as constant $\epsilon$-greedy. However, the latter may fall short of defining an optimal policy as it keeps the $\epsilon$ probability constant over time, resulting in the same amount of exploration throughout the process. Indeed, in the initial stages of learning, the agent should require more exploration than in the final stages due to the agent's greater uncertainty. To address this, we introduce an alternative deterministic approach where $\epsilon$ is modeled as a linear equation, the so-called Decreasing $\epsilon$-greedy, allowing it to adapt based on the evolving dynamics of the process \citep{DQN_conference}. This method adjusts the decrease rate for $\epsilon$, tailoring it differently for each number of steps. The system of the equations describing $\epsilon$ is given by
\begin{gather}
    \epsilon_t = \begin{cases}1, & \mbox{if }X_{steps} \le n_{start}\\ \delta_0 + \delta_1 \cdot X_{steps}, & \mbox{if } X_{steps} > n_{start} + \epsilon_{ann} \\ \pi_0 + \pi_1 \cdot X_{steps}, & \mbox{otherwise} \end{cases} \mbox{ ,}\label{eq:eps_decr}
\end{gather}
where $X_{steps}$ represents the current step number, $\pi_0$, $\pi_1$ are the intercepts and $\delta_0$, $\delta_1$ are the slopes of the two linear functions. These coefficients are derived from the following equations
\begin{gather}
    \pi_0 = \epsilon_{start} - \pi_1 \cdot n_{start} \mbox{ ,}\label{eq:beta0}\\
    \pi_1 = - \frac{\epsilon_{start} - \epsilon_{last}}{\epsilon_{ann}}\mbox{ ,}\label{eq:beta1}\\
    \delta_0 = \epsilon_{end} - \delta_1 \cdot n_{max}\mbox{ ,}\label{eq:delta0}\\
    \delta_1 = - \frac{\epsilon_{last} - \epsilon_{end}}{n_{max} -\epsilon_{ann} -n_{start}}\mbox{ .}\label{eq:delta1}
\end{gather}

The $\epsilon$ probability is formulated using two different equations to get a different balance of the exploration-exploitation trade-off. The change point is built upon $\epsilon_{ann}$ parameter (expressed in frames), which specifies the number of steps after which $\epsilon$ should decrease more slowly. This method is indeed characterized by two linear trends: the initial phase is governed by a steeper decrease with higher values of $\epsilon$, while the final phase adopts a more gradual trend with lower values of $\epsilon$.

To gather sufficient experience memory, all exploration strategies are modified to ensure an initial phase of full exploration. The parameter $n_{start}$ represents the upper limit in terms of steps that define the end of full exploration.

\subsection{Value-Difference Based Exploration}
The Value Difference Based Exploration (VDBE) strategy, introduced by \citet{VDBE} in Q-learning scenario, revolves around adjusting the $\epsilon$ probability based on Temporal Difference (TD) errors computed during the learning process. Unlike deterministic methods, VDBE is a data-driven $\epsilon$-greedy approach that dynamically updates $\epsilon$ using information gained from the agent's learning. In this work, We extend this method to a Deep Q-learning scenario and specifically to the recurrent case. We consider a value of $\epsilon$ to be updated for each epoch as the difference between the previous $\epsilon$ value and a new piece of information obtained by the Boltzmann distribution function, as
\begin{gather}
f(h,a,\nu) = \frac{1 - e^{- \frac{\Delta_{err}}{\nu}}}{1 + e^{- \frac{\Delta_{err}}{\nu}}}\mbox{ , and}\\
\epsilon_{t+1} = \lambda\cdot f(h,a,\nu) + (1-\lambda)\cdot\epsilon_t \mbox{ ,}\label{eq:eps_vdbe}
\end{gather}
where $\lambda$ and $\nu$ respectively consist of the weights of the selected action and the inverse of sensitivity, which always expresses a positive constant. We define $\Delta_{err}$ as the difference between the optimal Q-values obtained after and before the update, such that
\begin{gather}
    \Delta_{err}= Q_p(h_t, a^*) - Q_{p-1}(h_t, a^*) \mbox{ ,}\label{eq:td_error_drqn}
\end{gather}
where $p= (0,1, \dots, n)$ are the n-epochs of the neural network and $a^*$ is the optimal action chosen at the current step. 

\subsection{Bayesian Model Combination}
Similar to VDBE, $\epsilon$-Bayesian Model Combination (BMC) is a data-driven $\epsilon$-greedy strategy that dynamically updates the $\epsilon$ value based on agent's data acquired during the RL process. The approach, proposed by \citet{BMC}, adopts a Bayesian perspective instead of relying on a heuristic approach, as seen in VDBE, for parameter tuning. $\epsilon$-BMC leverages the strengths of Bayesian Model Combination, allowing a weighted combination of models. Specifically, such weights correspond to $\epsilon$ and $(1-\epsilon)$, as follows
\begin{gather}
    \tilde{G}= (1-\epsilon)\cdot\tilde{G}^Q + \epsilon\cdot\tilde{G}^U \mbox{ ,} \label{eq:bmc_model}
\end{gather}
where $\tilde{G}^Q$ e $\tilde{G}^U$ are, respectively, the two different models practiced in $\epsilon$-greedy strategy: greedy and uniform, represented by
\begin{gather}
    \tilde{G}^Q= r_{t+1} + \gamma \max_{a' \in A} Q_t(h_{t+1}, a') \mbox{ ,}\label{eq:bmc_greedy}\\
    \tilde{G}^U= r_{t+1} + \gamma \frac{1}{|A|} \sum_{a' \in A} Q_t(h_{t+1}, a')\mbox{ .}\label{eq:bmc_uniform}
\end{gather}
\citet{BMC} emphasized the distinction between the $\epsilon$-BMC strategy and Bayesian Q-learning, wherein Q-values are modeled using Normal-Gamma priors. $\epsilon$-BMC strategy concentrates solely on the variance distribution, employing Normal-Gamma prior. This method offers several advantages, including theoretical convergence guarantees within Q-learning. Furthermore, it proves to be a robust and efficient strategy for providing a good exploration balance. 

Assuming a Normal distribution for return observations $Q_{h,a}$, expressed as
\begin{gather}
    Q_{h,a}|m,\tau \sim N(\tilde{G}^{m}_t, \tau^{-1}) \mbox{ ,}\label{distr:qvalues}
\end{gather}
where $m$ denotes the model chosen (greedy or uniform) and $\tau^{-1}$ represents the precision, alongside adopting a Normal-Gamma model for these parameters, results in the derivation of a Normal-Gamma posterior distribution. As mentioned earlier, this approach specifically centers on the prior distribution of return variance. By marginalizing the Normal-Gamma posterior, the resulting marginal distribution of $\tau$, with parameters $a_t$ and $b_t$, corresponds to
\begin{gather}
    \tau|D \sim Gamma(a_t, b_t)\mbox{ , and}\label{distr:tau}\\
    a_t = a_0 + \frac{t}{2}, \hspace{3mm}
    b_t = b_0 + \frac{t}{2}\bigl(\tilde{\sigma}^2_t + \frac{\tau_0}{\tau_0+ t}(\hat{\mu}_t - \mu_0)^2\bigl)\mbox{.}
\end{gather}
The parameters $\hat{\mu}_t$ and $\tilde{\sigma}^2_t$ are calculated as the mean and the variance of the data of previously observed returns $D=\{Q_{h_i,a_i,i}, i= 1, \ldots, t-1 \}$ \cite{BMC}. Consequently, in order to obtain the distribution of $Q_{h,a}|m,D$ by marginalizing over $\tau$, it is possible to obtain t-distributed likelihood function as
\begin{gather}
    P(Q_{h,a}|m, D) =  \int_0^\infty P(Q_{h,a}|m,\tau) P(\tau|D) \,d\tau \mbox{ .}\label{eq:t-distr_q}
\end{gather}
Solving Eq. \eqref{eq:t-distr_q}, the kernel of a T-Student with three parameters $(\tilde{G}^m_t, \frac{a_t}{b_t}, 2a_t)$ is identified. Given the last result in Eq. \eqref{eq:t-distr_q}, \citet{BMC} computed the equation for obtaining the expected return $\mathbb{E}[Q_{h,a}|D]$ which involves the posterior distribution of $w|D$. This distribution characterizes the weights assigned to the greedy \eqref{eq:bmc_greedy} and uniform model \eqref{eq:bmc_uniform}. The expected return is given by
\begin{gather}
    \mathbb{E}[Q_{h,a}|D]=\int_0^1 \mathbb{E}[Q_{h,a}|w, D] \mathbb{P}(w|D) \,dw \mbox{ .}\label{eq:posterior_2}
\end{gather}
Eq. \eqref{eq:posterior_2} combines the estimates of Q-values, derived from the Double Dueling Deep Recurrent Q-Network model and the uniform model, and the weights, which could be used to find a value for $\epsilon$. Consequently, Eq. \eqref{eq:posterior_2} can be reformulated as
\begin{gather}
    \mathbb{E}[Q_{h,a}|D]= (1-\mathbb{E}[w|D]) \tilde{G}^Q_t  + \mathbb{E}[w|D] \tilde{G}^U_t \mbox{ .}\label{dist: posterior_qvalues}
\end{gather}
Subsequently, the study proceeds to determine the posterior distribution of the weights to establish a rule for updating $\epsilon$. Since the Eq. \eqref{dist: posterior_qvalues}, $\epsilon$ can be expressed as $\epsilon_t= \mathbb{E}[w|D]$. Based on this result, \citet{BMC} used a technique (\textit{i.e.} Dirichlet Moment-Matching) to approximate the posterior distribution for finding out the expected value of $w|D$, such that
\begin{gather}
\epsilon_t^{BMC} \approx E_{Beta(\alpha_t, \beta_t)}[w|D] = \frac{\alpha_t}{\alpha_t + \beta_t} \mbox{ .}\label{eq:eps_bmc}
\end{gather}
Beta parameters (i.e. $\alpha_t$ and $\beta_t$) are obtained by the following system of equations
\begin{gather}
    m_t = \frac{\alpha_t}{\alpha_t + \beta_t + 1} \frac{e_t^U (\alpha_t +1) + e_t^Q \beta_t}{e_t^U \alpha_t + e_t^Q \beta_t}\mbox{ ,}\\
    v_t = \frac{\alpha_t}{\alpha_t + \beta_t + 1} \frac{\alpha_t +1}{\alpha_t + \beta_t + 2} \frac{e_t^U (\alpha_t +2) + e_t^Q \beta_t}{e_t^U \alpha_t + e_t^Q \beta_t}\mbox{ ,}\\
    r_t = \frac{m_t - v_t}{v_t - m_t^2}\mbox{ ,}\\
    \alpha_{t+1} = m_t *r_t\mbox{ ,}\\
    \beta_{t+1} = (1-m_t) * r_t\mbox{ ,}
\end{gather}
where $e_t^U$ and $e_t^Q$ are the evidence of a return under the distribution of $Q_{h,a}|m,D$ \cite{BMC}.

\subsection{Max-Boltzmann Exploration}

The fundamental idea behind Max-Boltzmann Exploration (MBE) is to leverage the Softmax policy for exploring actions, so as to balance exploration-exploitation trade-off with $\epsilon$-greedy to take advantage of the strengths of both strategies. This approach was proposed by \citet{mbe} and it integrates $\epsilon$-greedy strategy with a difference in action sampling in the exploration phase, where actions are sampled over Boltzmann probability distribution as in Softmax policy. Indeed, at each time step, the action can be selected according to the following rule
\begin{gather}
    a_t=\begin{cases}\argmax_a Q_t(h_t,a), & \mbox{with probability } 1-\epsilon \\ \mbox{Softmax policy}, & \mbox{with probability } \epsilon \end{cases} \mbox{ .}\label{eq:mbe}
\end{gather}
The Max-Boltzmann policy effectively addresses a limitation of $\epsilon$-greedy concerning the exploration probability distribution. The classical $\epsilon$-greedy exploration assigns equal probabilities to all actions, thereby assigning too large probabilities to explore worse actions \cite{mbe}. In contrast, including a Softmax policy into the $\epsilon$-greedy method, can establish a probability distribution that deviates significantly from a uniform distribution, thereby avoiding a substantial decline in performance. The degree of separability from the uniform distribution is tied to the temperature parameter: with a high $\tau$, the Softmax policy will converge to a random policy. On the other hand, Max-Boltzmann Exploration considers a probability of exploration by $\epsilon$ that overcomes the Softmax method, providing the flexibility to choose the extent of exploration.

An improvement of MBE could be drafted by VDBE-Softmax, wherein the $\epsilon$ probability is adjusted using the VDBE method, thus resulting in a heuristic strategy grounded in a data-driven approach \cite{VDBE_softmax}. The $\epsilon$ update is regulated like Eq. \eqref{eq:eps_vdbe}, totally integrated in MBE framework.

\section{Experimental settings and simulation platform} \label{sec:exp}

In this section, we describe the experimental settings of our proposed method using AirSim as a simulation platform for autonomous driving. Specifically, our emphasis is on evaluating exploration strategies with both deterministic and adaptive approaches while controlling the performance in terms of collision avoidance. To conduct our analysis, at the beginning of the process, we position the car at specific points in the environment by randomly drawing out the coordinates from a predefined set of starting points. Hence, we assess the performance of RL agents in terms of learning, employing a set of ten training starting points.  Additionally, we evaluate the predictive capabilities using another set of ten starting points (test set). The RL processes involve 1 million steps and each episode ends when a collision event occurs. The experiments were executed on a virtual machine with two GPUs Tesla M60 and Linux OS. AirSim simulator has been connected to Python. TensorFlow Compact V2, OpenAI, and OpenCV packages were used. 


\subsection{AirSim simulator: Neighborhood Environment}\label{subsec:Airsim}

\begin{figure}[ht]
	\begin{center}
	    \includegraphics[scale=1.1]{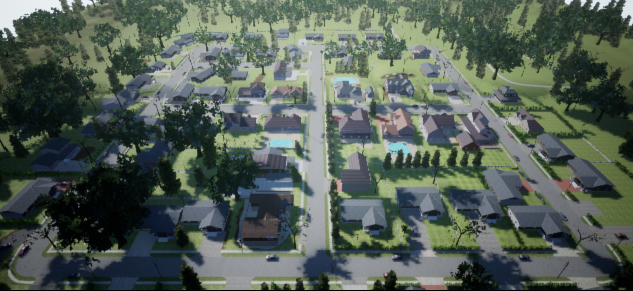}
	    \caption[Image]{\textbf{AirSim NH environment.} The Neighborhood Environment is shown through the AirSim simulation platform.}
	    \label{fig:airsimnh}
	\end{center}
\end{figure}

As previously mentioned, we use AirSim simulator to train and test Deep Double Dueling Recurrent Q-Learning models for learning self-driving cars. AirSim is an open-source program based on Unreal Engine, which offers various environments with several landscapes. With the aim of learning models for collision avoidance, we chose the AirSim NH environment. This environment represents a small, simplified, urban neighbourhood with a rectangular shape. AirSim NH features a main street connected with side streets, including elements such as trees, parked cars, shadows, and lights, mimicking real-world scenarios (see Figure \ref{fig:airsimnh}). \citet{AirSim} provided an extensive analysis of AirSim highlighting its provision of realistic environments both in terms of physical and visual attributes. For executing Airsim, we employed OpenGL drivers and established a connection with Python. Within the environment, we set a single car, called PhysicXCar, along with specifying a single weather condition.

\subsection{Models experimental setup}\label{subsec:Experimentalsetup}

To facilitate a comprehensive comparison and evaluation of exploration strategies, we considered a predefined set of parameters based on \citet{DQN_conference} results.

\begin{table}[H]
	\caption{Hyper-parameters of Double Dueling Deep Recurrent Q-Network.}
	\label{tab:hyperparam_drqn}
	\begin{center}
        \begin{tabular}{llrrrr}
        \toprule
         Hyper-parameter   & Range & Unit\\
        \midrule
        Buffer size & $\{1,000; 2,000\}$ & \emph{Episode}\\
        Batch size & $10$ & \emph{Episode}\\
        Update rate & $4$& \emph{Step}\\
        Trace length & $10$& \emph{Step}\\
        Error masked & $7$& \emph{Step}\\
        State updated & $3$& \emph{Step}\\
        Starting update & $999$& \emph{Episode}\\
        End process & $1,000,000$ & \emph{Step}\\
        \bottomrule
        \end{tabular}
	\end{center}
\end{table}
In Table~\ref{tab:hyperparam_drqn} the parameter settings for the Double Dueling Deep Recurrent Q-Network algorithm\footnote{The code is available at \url{https://github.com/ValentinaZangirolami/DRL}.} are shown.
We varied the buffer size with two assigned values to assess the impact of experience buffer size changes on neural network optimization. Additionally, we opted for a soft update for the Target Network, where, at each updating step, the weights of the Target Network are updated with 0.1\% of the main network weights. Furthermore, we initiated a completely random process for the first \emph{Starting update} episodes.

\subsubsection{Exploration strategies}\label{subsubsec:Experimentalsetup-ee}

We tested seven different exploration strategies, as detailed in Section~\ref{sec:expl_str}, with a singular setting for each strategy. Table~\ref{tab:hyperparam_ee} shows all parameter settings used in the respective fourteen tests.
\begin{table}[H]
	\caption{Hyper-parameters of Exploration strategies.}
	\label{tab:hyperparam_ee}
	\begin{center}
        \begin{tabular}{llrrrr}
        \toprule
         Strategy & Hyper-parameter   & Value \\
        \midrule
        \emph{Constant, MBE} &$\epsilon$ & $0.05$ \\
        \emph{Decreasing $\epsilon$-greedy}&$\epsilon_{start}$ & 1\\
        \emph{Decreasing $\epsilon$-greedy}&$\epsilon_{last}$ & 0.1\\
        \emph{Decreasing $\epsilon$-greedy}&$\epsilon_{end}$ & 0.01\\
        \emph{VDBE, VDBE-Softmax}& $\nu$& $1$ \\
        \emph{VDBE, VDBE-Softmax}& $\lambda$& $0.2$ \\
        \emph{BMC}& $\alpha_0, \beta_0$ & 25 \\
        \emph{BMC}& $a_0, b_0$ & 250 \\
        \emph{BMC}& $\mu_0$ & 0 \\
        \emph{BMC}& $\tau_0$ & 1 \\
        \emph{Softmax, MBE, VDBE-Softmax}& $\kappa$ & 0.1 \\
        \bottomrule
        \end{tabular}
	\end{center}
\end{table}

In particular, the parameters of BMC method align with those considered by \citet{BMC} in their experimental settings. Similarly, the parameters of VDBE and VDBE-Softmax are based on experiments carried out by \citet{VDBE} and \citet{VDBE_softmax}. While considering an exploration strategy involving an update regulated by the VDBE equation, the $\lambda$ parameter was set equal to the inverse of the number of actions, and $\nu$  was assigned to a value within the range, avoiding extremes, as recommended by the authors. For Softmax, we assigned a moderate value for $\kappa$, following the results obtained by \citet{VDBE_softmax}. Consequently, we extended this value to MBE and VDBE-Softmax. Differently, the $\epsilon$ value assumed in constant $\epsilon$-greedy is based on the values derived from the results of stochastic $\epsilon$-greedy methods. The Decreasing $\epsilon$-greedy method required a change in the updated structure of D3RQN model, since dependents on steps instead of episodes. Specifically, in Table~\ref{tab:hyperparam_drqn} we set the starting point of the update to 999 episodes. In Decreasing $\epsilon$-greedy, this parameter is adjusted in steps format ($n_{start}$) in Eq. \eqref{eq:beta0} and Eq. \eqref{eq:delta1}. We assigned a value of 50,000 steps to the $n_{start}$ parameter and 400,000 steps for $\epsilon_{ann}$. The remaining parameters of the Decreasing $\epsilon$-greedy strategy are based on \citet{DQN_conference}.

Furthermore, we standardized the management of $\epsilon$ across all methods. According to the Decreasing $\epsilon$-greedy strategy, we set $\epsilon$ equal to 1 to fill the experience buffer until the agent update starts.

\subsection{Results and comparison}\label{sec:results}

In this section, we present the results derived from the fourteen Double Dueling Deep Recurrent Q-Network models (as the settings outlined in Section \ref{sec:exp}) using the AirSim simulator, to evaluate the impact of the seven exploration strategies and the two buffer capacities on agent learning.

\begin{figure}[ht]
	\begin{center}
	    \includegraphics[scale=0.87]{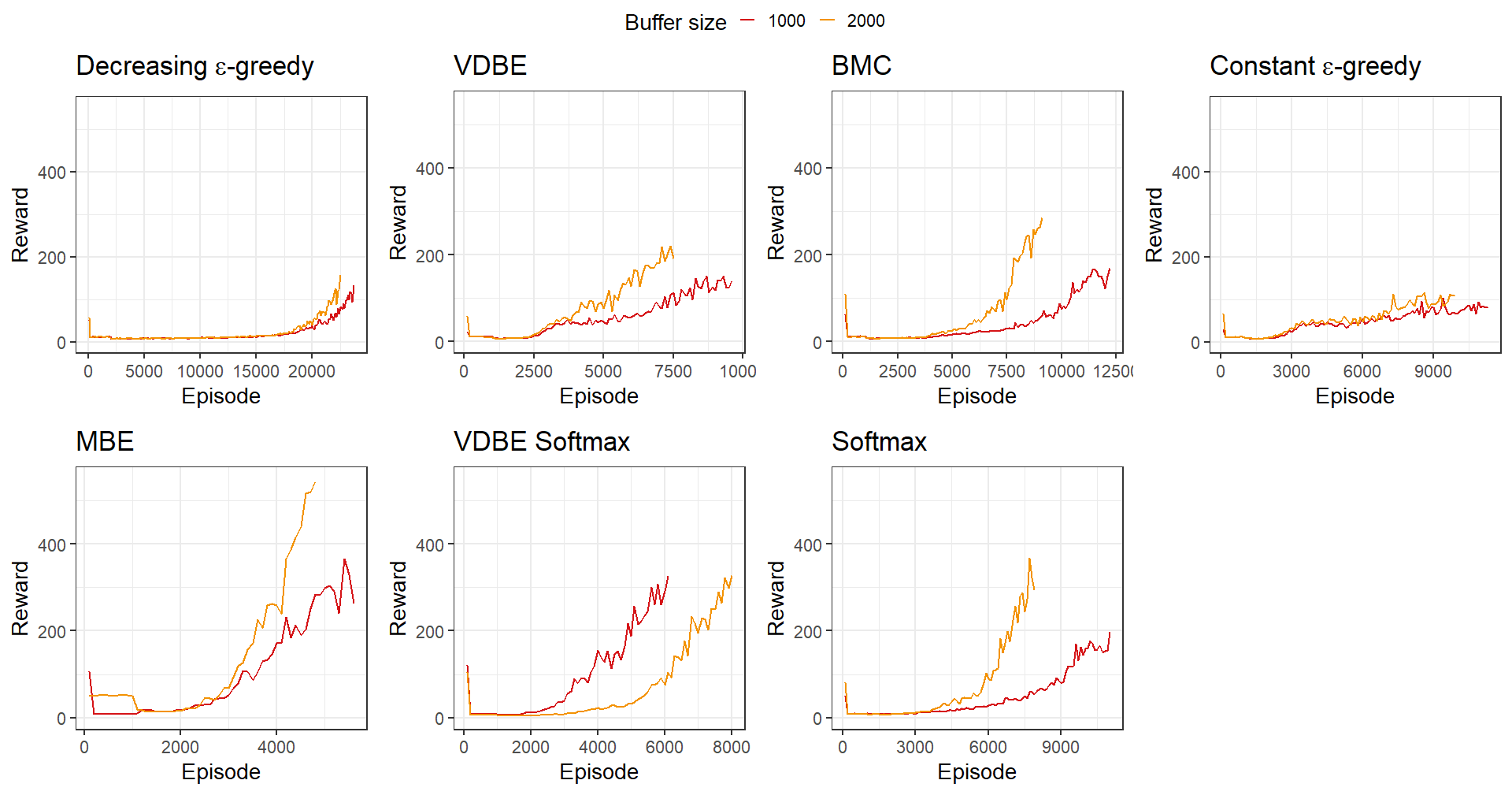}
	    \caption[Reward]{\textbf{Comparison of the exploration strategies with D3RQN agent.} The training curves show the average reward per 100 episodes for each value of the buffer size. The horizontal axis and the vertical axis indicate, respectively, the number of episodes and the average reward.}
	    \label{fig:reward_train}
	\end{center}
\end{figure}

Figure~\ref{fig:reward_train} illustrates the trends in terms of Reward values during the training process. All models exhibit an increasing trend, indicative of effective policy learning. However, deterministic strategies reach lower average reward than other strategies. An intriguing observation is the notable contrast between $\epsilon$-greedy strategies and those employing the Softmax method for exploration, such as Softmax, MBE, and VDBE-Softmax, which perform exceptionally well, reaching significantly higher values in the final steps. This effect might be attributed to the use of a different distribution function for exploration. As discussed in Section~\ref{sec:expl_str}, the $\epsilon$-greedy random policy may induce wrong actions, thereby leading to a substantial worsening of performance and, thus, the results obtained from the training may be misleading.

\begin{figure}[ht]
  \begin{subfigure}{0.31\textwidth}
  \captionsetup{justification=centering}
    \includegraphics[width=\linewidth]{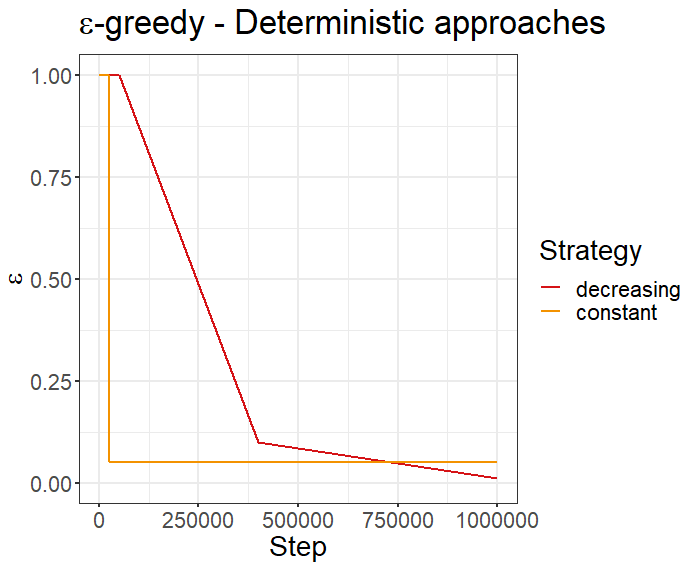} \caption{}\label{fig:eps_a}
  \end{subfigure}%
  \hspace*{\fill}
  \begin{subfigure}{0.31\textwidth}
  \captionsetup{justification=centering}
    \includegraphics[width=\linewidth]{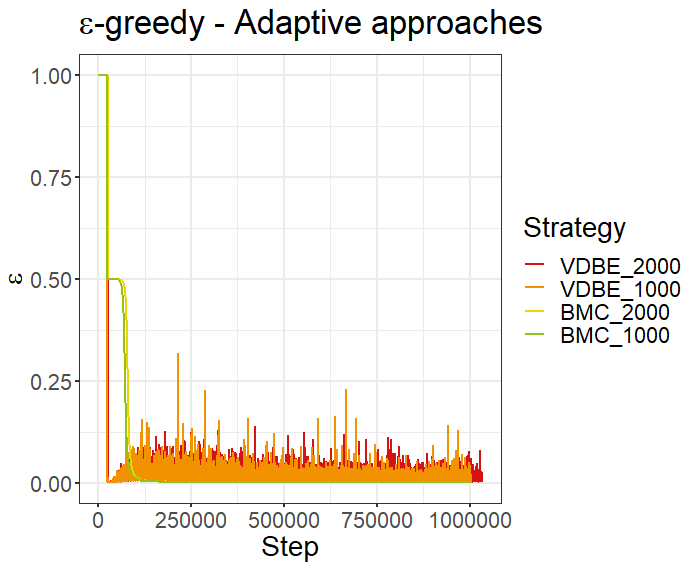}
    \caption{}\label{fig:eps_b}
  \end{subfigure}%
  \hspace*{\fill}
  \begin{subfigure}{0.31\textwidth}
  \captionsetup{justification=centering}
    \includegraphics[width=\linewidth]{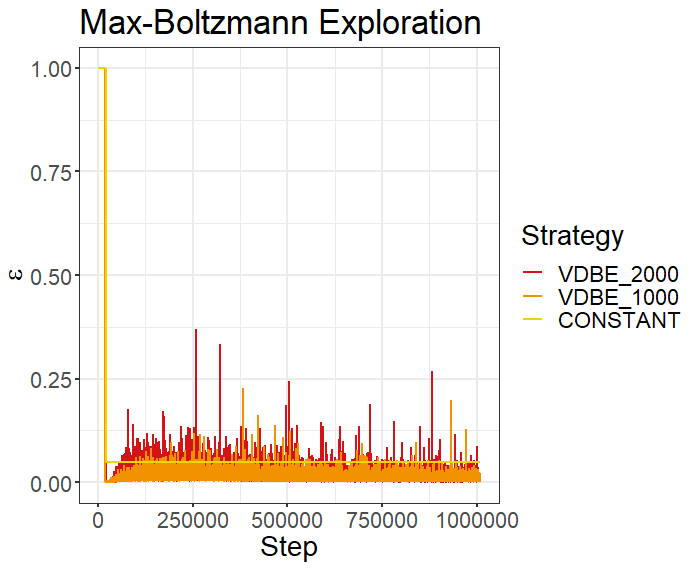} \caption{}\label{fig:eps_c}
  \end{subfigure}

\caption[Epsilon]{\textbf{Comparison of $\epsilon$ values for $\epsilon$-greedy and Max-Boltzmann methods.} The horizontal axis and the vertical axis indicate, respectively, the number of environment steps and the $\epsilon$ value. Figure \textbf{(a)} shows $\epsilon$ values per steps for deterministic $\epsilon$-greedy strategies. Figure \textbf{(b)} shows $\epsilon$ values per step for adaptive $\epsilon$-greedy strategies (BMC and VDBE) distinguished by buffer size. Figure \textbf{(c)} shows $\epsilon$ values per step for Max-Boltzmann Exploration methods (constant and VDBE) distinguished by buffer size.} \label{fig:epsilon_train}
\end{figure}

As illustrated in Figure~\ref{fig:epsilon_train}, stochastic strategies for $\epsilon$-greedy exhibit very low values, even close to zero, whereas the deterministic method consists of much higher values throughout the learning process. Despite the VDBE-Softmax strategy being characterised by a Boltzmann distribution function with $\tau$ equal to 0.1, thus being very far from uniform distribution, the oscillation of $\epsilon$ is very similar to VDBE.

\begin{figure}[ht]
	\begin{center}
	    \includegraphics[scale=0.85]{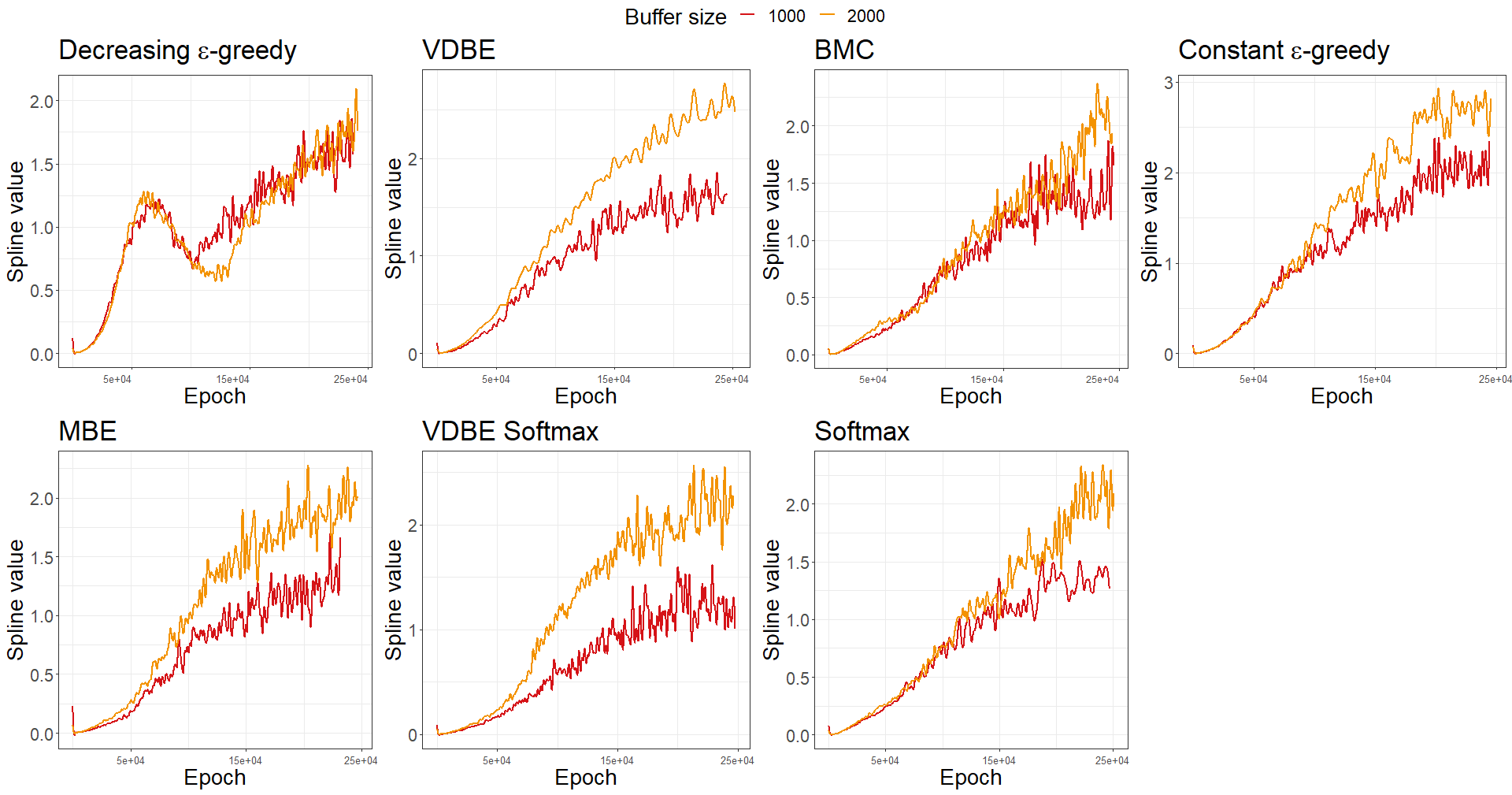}
	    \caption[Loss]{\textbf{Comparison of training loss differences for each exploration strategy and buffer size.} The vertical axis denotes spline values of the absolute difference between the previous and the next loss. The horizontal axis indicates the training epochs.}
	    \label{fig:loss_train}
	\end{center}
\end{figure}

In Figure~\ref{fig:loss_train}, we analyze the spline of the loss differences for each model. It should be noted that, in Deep Reinforcement Learning tasks, the loss function is based on the difference between target values and prediction values. It can be generally observed that the gap between loss values before and after the update is too high with a larger size of buffer of experience. Furthermore, loss differences are too high when the agent improves performance in the long term.

In order to evaluate the goodness of steering angle prediction, we reported a summary of the performance gained from 300 episodes that were evaluated on 10 training starting points and on 10 test starting points respectively. We used different statistical indicators (among others, average, standard deviation, and minimum) of episode length and Collision-Free Rate (CFR) to evaluate the performance. 

\begin{table}[H]
	\caption{\textbf{Training set evaluation.} Performance of the agent over the training set with a buffer size equal to 2,000. The agent was evaluated with 30 trials for each of the 10 training starting points. The agent performance is based on summary statistics of episode length and the collision-free rate (CFR).}
	\label{tab:train_results_bs2000}
	\begin{center}
        \begin{tabular}{l|c|c|c|c}
        \toprule
         Strategy & Average & Standard Deviation & Min & CFR \\
        \midrule
        \emph{Decreasing $\epsilon$-greedy}& 1,452.02 & 688.11 & 29 & 53.36\%\\
        \emph{Constant $\epsilon$-greedy}& 1,317.61 & 736.28 & 64 & 44.41\%\\
        \emph{VDBE}& 1,795.24 & 474.16 & \textbf{86} & 80\%\\
        \emph{BMC}& 1,769.24 & 532.5 & 34 & 80.47\%\\
        \emph{Softmax}& 1,802.86 & 487.04& 43 & 81.69\%\\
        \emph{VDBE-Softmax}& \textbf{1,917.26} & \textbf{325.36} & 81 & \textbf{91.92}\%\\
        \emph{MBE}& 1,798.32 & 488.14 & 79 & 81.54\%\\
        \bottomrule
        \end{tabular}
	\end{center}
\end{table}

The results illustrated in Tables~\ref{tab:train_results_bs2000} and~\ref{tab:train_results_bs1000} show that all strategies, except VDBE, perform better with a smaller buffer size. In general, Collision-free rate and the average of steps have higher values together with a reduction in episode length variability, especially for the Softmax, VDBE-Softmax, MBE and $\epsilon$-greedy BMC methods. 

\begin{table}[H]
	\caption{\textbf{Training set evaluation.} Performance of the agent over the training set with a buffer size equal to 1,000. The agent was evaluated with 30 trials for each of the 10 training starting points. The agent performance is based on summary statistics of episode length and the collision-free rate (CFR).}
	\label{tab:train_results_bs1000}
	\begin{center}
        \begin{tabular}{l|c|c|c|c}
        \toprule
         Strategy & Average & Standard Deviation & Min & CFR \\
        \midrule
        \emph{Decreasing $\epsilon$-greedy}& 1,465.30 & 690.08 & 46 & 53.66\%\\
        \emph{Constant $\epsilon$-greedy}& 1,608.84 & 612.48 & \textbf{114} & 62\%\\
        \emph{VDBE}& 1,728.66 & 565.27 & 71 & 76.09\%\\
        \emph{BMC}& \textbf{1,973.01}& \textbf{219.75} & 7 & \textbf{97.62}\%\\
        \emph{Softmax}& 1,942.69 & 285.46 & 16 & 94.59\%\\
        \emph{VDBE-Softmax}& 1,949.84 & 259.87 & 14 & 95.31\%\\
        \emph{MBE}& 1,947.92 & 243.59 & 17 & 94.28\%\\
        \bottomrule
        \end{tabular}
	\end{center}
\end{table}

 In Tables~\ref{tab:test_results_bs2000} and~\ref{tab:test_results_bs1000}, the results obtained by 300 episodes for each model are exhibited. Since a single initial car position may not be sufficient to represent the performance and the generalization, ten different starting points are considered during testing.  
 
\begin{table}[H]
	\caption{\textbf{Test set evaluation.} Performance of the agent over the test set with a buffer size equal to 2,000. The agent was evaluated with 30 trials for each of the 10 test starting points. The agent performance is based on summary statistics of episode length and the collision-free rate (CFR).}
	\label{tab:test_results_bs2000}
	\begin{center}
        \begin{tabular}{l|c|c|c|c}
        \toprule
         Strategy & Average & Standard Deviation & Min & CFR \\
        \midrule
        \emph{Decreasing $\epsilon$-greedy}& 1,232.26 & 818.09 & 69 & 44.48\%\\
        \emph{Constant $\epsilon$-greedy}& 1,240.52 & 744.12 & 41 & 35.73\%\\
        \emph{VDBE}& 1,593.84 & 700.01 & 40 & 68.64\%\\
        \emph{BMC}& 1,604.47 & 718.93 & 40 & 72.47\%\\
        \emph{Softmax}& \textbf{1,666.6} & \textbf{655.72}& \textbf{92} & \textbf{75.92}\%\\
        \emph{VDBE-Softmax}& 1,609.05 & 749.3 & 84 & 75.85\%\\
        \emph{MBE}& 1,561.56 & 731.04 & \textbf{92} & 69.31\%\\
        \bottomrule
        \end{tabular}
	\end{center}
\end{table}

\begin{table}[H]
	\caption{\textbf{Test set evaluation.} Performance of the agent over the test set with buffer size equal to 1,000. The agent was evaluated with 30 trials for each of the 10 test starting points. The agent performance is based on summary statistics of episode length and the collision-free rate (CFR).}
	\label{tab:test_results_bs1000}
	\begin{center}
        \begin{tabular}{l|c|c|c|c}
        \toprule
         Strategy & Average & Standard Deviation & Min & CFR \\
        \midrule
        \emph{Decreasing $\epsilon$-greedy}& 1,316.54 & 789.73 & 59 & 48.95\%\\
        \emph{Constant $\epsilon$-greedy}& 1,508.11 & 722.97 & 29 & 59.79\%\\
        \emph{VDBE}& 1,483.31 & 755.27 & 87 & 61.77\%\\
        \emph{BMC}& \textbf{1,821.72}& \textbf{509.57} & 86 & 86.71\%\\
        \emph{Softmax}& 1,724.11 & 641.66 & 48 & 82.25\%\\
        \emph{VDBE-Softmax}& 1,791.01& 573.97 & \textbf{90} & \textbf{87.54}\%\\
        \emph{MBE}& 1,738.52 & 636.68 & 87 & 84.14\%\\
        \bottomrule
        \end{tabular}
	\end{center}
\end{table}

\begin{figure}[ht]
	\begin{center}
	    \includegraphics[scale=1.2]{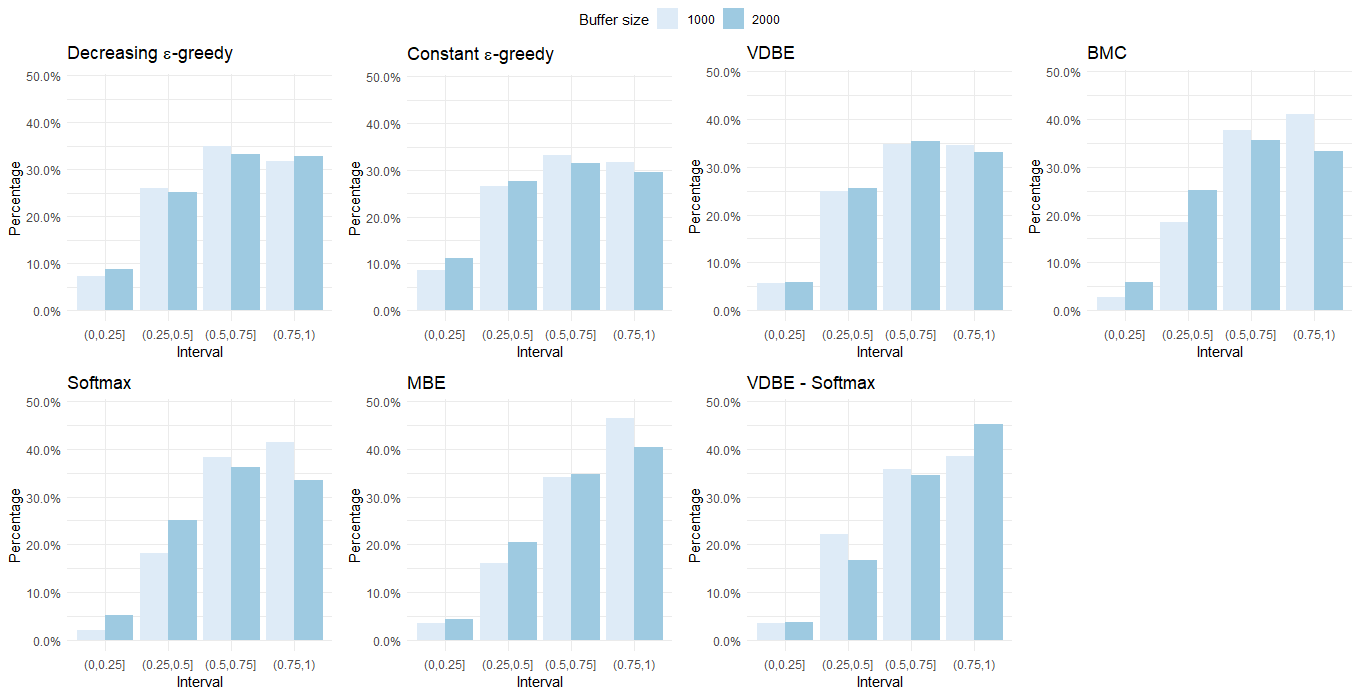}
	    \caption[Reward Test]{\textbf{Comparison of test reward values for each exploration strategy and buffer size.} Each barplot shows the percentage of reward values falling within each range. The horizontal axis indicates different intervals of reward values in ascending order. The last bin contains the maximum values of reward meaning the optimal choices.}
	    \label{fig:reward_test}
	\end{center}
\end{figure}

We observed that performance remains very high on the test set, with a tiny discrepancy in comparison with the training set. The only drawback lies in the increasing step variability, which may suggest less stability. Generally, Softmax and related methods have better performance than $\epsilon$-greedy strategies, except for the BMC method. Finally, in Figure~\ref{fig:reward_test} we analyzed the optimality related to the action choice in order to understand which model was able to maximise the reward function at each step. 

In general, the strategies that achieve the best results above involve more optimal actions. BMC, Softmax, MBE, and VDBE-Softmax strategies have a higher frequency of observations in the last bin, which contains the highest reward values. Furthermore, almost all of the methods have higher frequencies in the last bins when the buffer size equals 1,000. Particularly, Max-Boltzmann Exploration is the strategy that is most successful in terms of optimal actions, as almost 50\% of the reward values are within the range [0.75,1].

\section{Conclusions and future work}\label{sec:conclusion}

This work investigates exploration strategies within a recurrent DRL framework. The problems of uncertainty and partial observability when applying Deep Reinforcement Learning are studied in the autonomous driving context.  Recurrent neural networks often exhibit slow convergence bringing low learning efficiency within the DRL scenario. In an attempt to mitigate this issue, we combined the D3RQN model, featuring a Convolutional Recurrent Neural Network for estimating Q-values related to five steering angles, with an adjustment of the LSTM layer to speed up the learning of the neural network in the training process. Subsequently, several exploration strategies were evaluated to address the exploration-exploitation dilemma, employing both deterministic and adaptive approaches, among other $\epsilon$-greedy and Softmax. While $\epsilon$-greedy, with its popular strategy of decreasing $\epsilon$ over time, has been extensively studied in autonomous driving scenarios, Softmax and adaptive methods remain less explored.

Experimental results under deterministic approaches showed Softmax and MBE outperformed $\epsilon$-greedy. However, adaptive strategies better balance exploration, enhancing collision avoidance. Notably, the uniform distribution used for sampling actions in $\epsilon$-greedy exploration appeared to impair the agent learning in the training process, leading to worsening performance. Despite that, using Bayesian inference for $\epsilon$-greedy updating yielded good results, even though the estimation of $\epsilon$ was close to zero after a few epochs and mitigating the potential negative impact of random sampling. In conclusion, the combination of Boltzmann distribution and $\epsilon$-greedy proved effective in leading exploration and facilitating learning in the environment, outperforming other methods and achieving more optimal actions.

Future research could continue to explore Max-Boltzmann exploration strategies. As proposed by \citet{BMC}, it would be interesting to incorporate Bayesian inference for estimating $\epsilon$ probability, as it is the case with $\epsilon$-greedy BMC method.



 \bibliographystyle{elsarticle-num-names} 
 \bibliography{literature}

\end{document}